\documentclass[10pt,journal,compsoc]{IEEEtran}
\ifCLASSOPTIONcompsoc
  \usepackage[nocompress]{cite}
\else
  \usepackage{cite}
\fi
\ifCLASSINFOpdf
\else
\fi

\usepackage{graphicx}
\usepackage{subfigure}
\usepackage{caption}
\usepackage{booktabs}

\usepackage{color}
\usepackage{amsmath}
\usepackage{amsthm}
\usepackage{amssymb}
\usepackage{algorithm}
\usepackage{algorithmic}
\usepackage[figuresright]{rotating}
\usepackage{arydshln}
\usepackage{multirow}

\begin{document}
%
\title{Test-Time Compensated Representation Learning for Extreme Traffic Forecasting}
%
%
%
%

\author{Zhiwei Zhang, 
        Weizhong Zhang, 
        Yaowei Huang,
        and~Kani Chen 
\IEEEcompsocitemizethanks{\IEEEcompsocthanksitem Zhiwei Zhang, Weizhong Zhang, Yaowei Huang and Kani Chen are with the Department
of Mathematics, The Hong Kong University of Science and Technology, Hong Kong,
China.\protect\\
E-mail: makchen@ust.hk
}
\thanks{Manuscript received September 15, 2023.}}
%
%

\markboth{Journal of \LaTeX\ Class Files,~Vol.~14, No.~8, August~2015}%
{Shell \MakeLowercase{\textit{et al.}}: Bare Demo of IEEEtran.cls for Computer Society Journals}
%



\IEEEtitleabstractindextext{%
\begin{abstract}
Traffic forecasting is a challenging task due to the complex spatio-temporal correlations among traffic series. In this paper, we identify an underexplored problem in multivariate traffic series prediction: extreme events. Road congestion and rush hours can result in low correlation in vehicle speeds at various intersections during adjacent time periods. Existing methods generally predict future series based on recent observations and entirely discard training data during the testing phase, rendering them unreliable for forecasting highly nonlinear multivariate time series. To tackle this issue, we propose a test-time compensated representation learning framework comprising a spatio-temporal decomposed data bank and a multi-head spatial transformer model (CompFormer). The former component explicitly separates all training data along the temporal dimension according to periodicity characteristics, while the latter component establishes a connection between recent observations and historical series in the data bank through a spatial attention matrix. This enables the CompFormer to transfer robust features to overcome anomalous events while using fewer computational resources. Our modules can be flexibly integrated with existing forecasting methods through end-to-end training, and we demonstrate their effectiveness on the METR-LA and PEMS-BAY benchmarks. Extensive experimental results show that our method is particularly important in extreme events, and can achieve significant improvements over six strong baselines, with an overall improvement of up to 28.2\%.
\end{abstract}

}

\maketitle

\IEEEdisplaynontitleabstractindextext

%
\IEEEpeerreviewmaketitle

\IEEEraisesectionheading{\section{Introduction}
\label{sec:introduction}}

%
%
%
%

\IEEEPARstart{W}{ith} the construction of smart cities, traffic speed forecasting plays an essential role in vehicle dispatching and route planning for intelligent transportation systems~\cite{review2021}. Recently, a significant amount of deep models have been subsequently developed to this research area, achieving noticeable improvements over traditional methods~\cite{dcrnn2017,graphwave2019,mtgnn2020,agcrn2020,sttn2020,t-transformer2020,stemm-gnn2020nips,dgcrn2021,dynamic2021,gts2021,dmstgcn2021kdd,st-norm2021,stgode2021kdd}. Especially graph neural network (GNN) based methods have attracted tremendous attention and demonstrated predictive performance due to their ability to capture the complicated and dynamic spatial correlations from graph~\cite{dcrnn2017,graphwave2019,mtgnn2020,agcrn2020,stemm-gnn2020nips,dgcrn2021,gts2021,dmstgcn2021kdd,stgode2021kdd}. However, multivariate traffic series forecasting remains challenging because it is difficult to simultaneously model complicated spatial dependencies and temporal dynamics, especially for the extreme events \cite{benchmark2021tkde}. Existing literature has paid little attention to extreme traffic forecasting.

\begin{figure*}
\centering
\subfigure[Two types of evaluation metrics]{\includegraphics[scale=0.535]{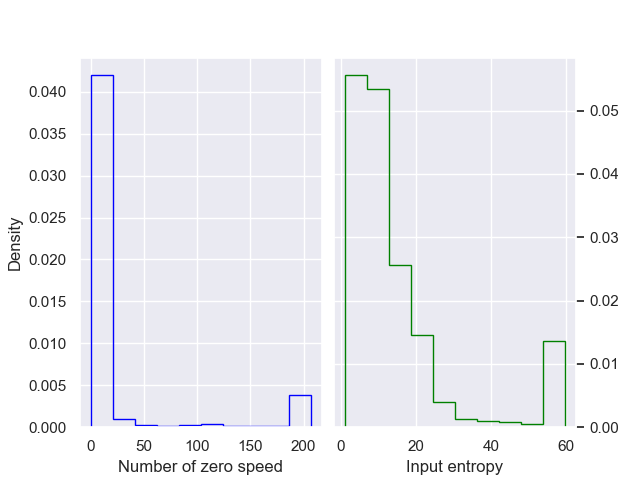}} \hspace{15pt}
\subfigure[Prediction errors under extreme events]{ \includegraphics[scale=0.535]{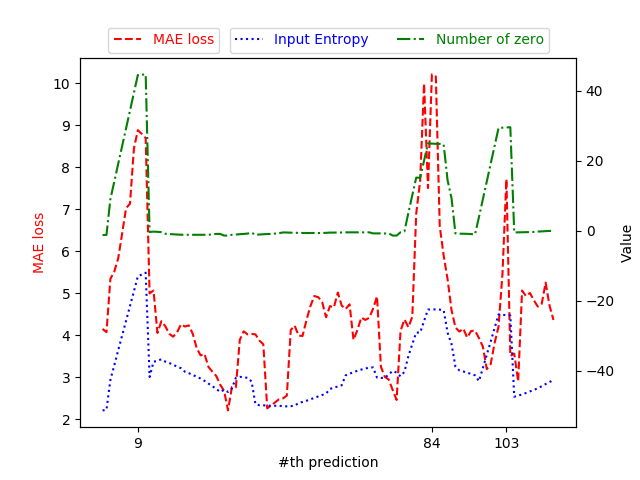}}
\caption{(a) Two types of evaluation metrics to measure the extreme events: one is the number of zero-valued observations, and the other is the entropy of input. This figure is made from METR-LA data. (b) The curves above show that the MAE loss achieved by GWN~\cite{graphwave2019} is always large when the entropy and number of zero-valued speed in input are large. The \textcolor{red}{left Y-axis} represents the MAE loss value, and the \textbf{right Y-axis} represents the values of input entropy and number of zero.}
\label{fig:extreme}
\end{figure*}

\begin{figure*}
\centering
\vspace{5pt}
\subfigure[Normal traffic conditions]{\includegraphics[scale=0.421]{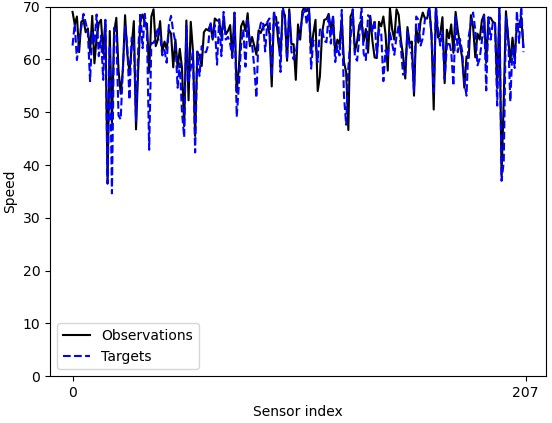}} \hspace{-1pt}
\subfigure[Eextreme event: zero-valued observations]{ \includegraphics[scale=0.421]{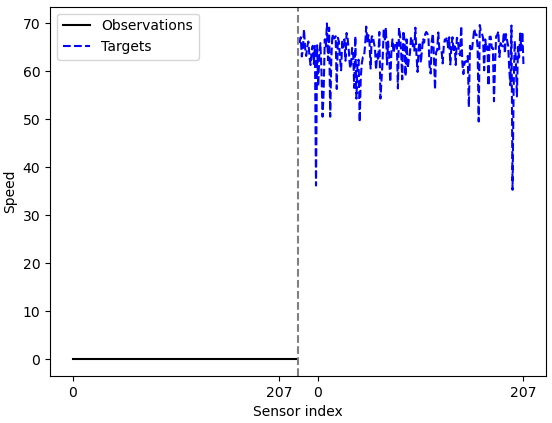}} \hspace{-1pt}
\subfigure[Eextreme event: traffic congestion]{ \includegraphics[scale=0.421]{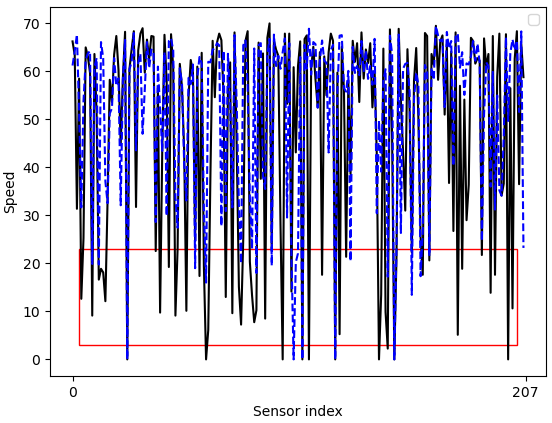}}
\caption{(a) Under normal traffic conditions, the vehicle speed at each intersection is stable at around 60 km/h, thus the prediction error will be very small. (b) In the first extreme case, the observed series and the predicted values are completely uncorrelated, which makes it difficult for the model to accurately predict. (c) In the second extreme case, traffic jams lead to huge differences in the vehicle speeds at various intersections, which also increase the difficulty of the model's predictions. The figures are made from METR-LA data.}
\label{fig:extreme-vis}
\end{figure*}

In this paper, we first identify extreme events in traffic speed forecasting, as shown in Figure~\ref{fig:extreme-vis}. An extreme event occurs when recent observations and forecasts exhibit different distributions, indicating a high degree of non-linearity between them, which increases the prediction difficulty of the model. We then define two evaluation metrics to measure extreme events: events with a large number of zero-valued speeds and events with high entropy in multivariate the traffic series. As illustrated in Figure~\ref{fig:extreme} (a), those two evaluation indicators exhibit a long-tailed distribution. More zero-valued speeds and larger input entropy indicate complex road conditions. We observe that existing methods struggle to handle extreme events well because they predict future series based on a limited horizon of recent observations. For example, they predict vehicle speeds for the next 30 minutes based on the observations collected in the last hour~\cite{dcrnn2017,graphwave2019,mtgnn2020}. The drawback of this approach is that limited-horizon observations can cause forecasting models to fail when faced with extreme events, as illustrated in Figure~\ref{fig:extreme} (b).
We argue that simply extending the field of view of observations is not a viable solution to this dilemma. Due to daily and weekly periodic characteristic and temporal dynamics, the simple enlarged long-distance historical data is redundant and unreliable. Additionally, learning informative patterns from long-term historical data is computationally expensive~\cite{step2022kdd}. Therefore, it is crucial to determine how to enlarge the horizon to the entire historical data within limited computing resource to improve predictions during extreme events.

\begin{figure*}[t] 
    \centering
    \vspace{3pt}
        \includegraphics[scale=0.675]{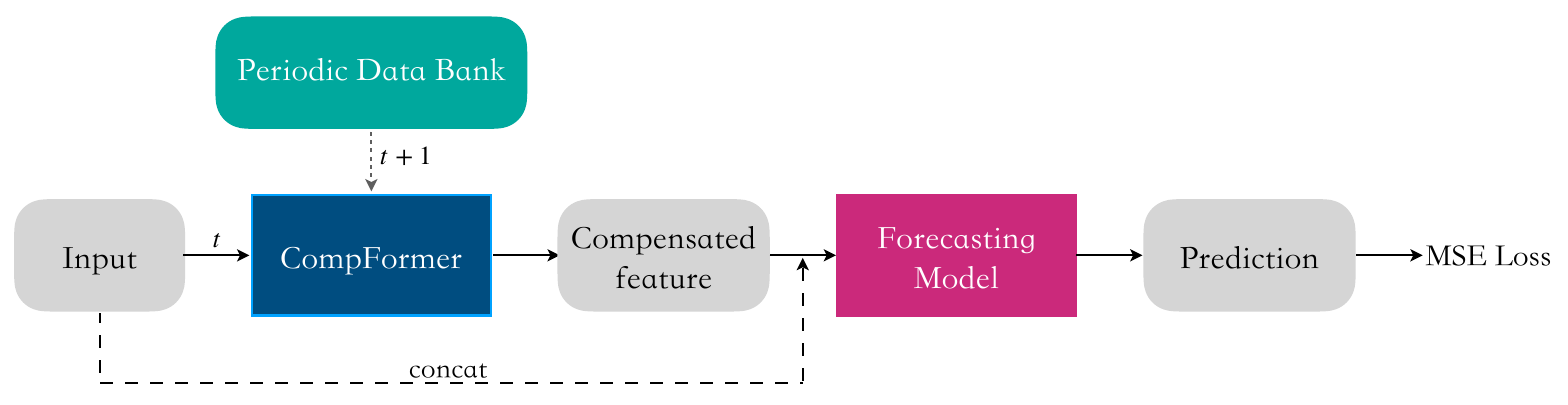}
    \caption{The overall framework of our proposed test-time compensation learning method. 1) All historical series are integrated into a periodic data bank. 2) The CompFormer model is used to establish a connection between recent observations and selected historical series using a spatial attention matrix to transfer robust representation for overcoming extreme events. 3) The learned compensated features are concatenated with the original embeddings as a new input for the forecasting models. The CompFormer and forcasting model can be trained end-to-end.}
    \label{fig:compformer}
\end{figure*}

We propose a test-time compensated representation learning framework to enhance model predictions under extreme events. Unlike other time series, such as stock prices, the historical traffic series have a longer shelf life in predicting the future as their patterns are always exhibit periodicity over time. Thus, the first component of our method divides all training data according to periodic characteristics (e.g., daily and weekly periods) and stores it in a data bank that can be retrieved based on the input timestamps. The spatio-temporal decomposed data bank explicitly separates all historical data in temporal dimension to make the following transformer model easy to access historical series with similar periods. The 3D visualization in Figure~\ref{fig:vis_pccs_3d} shows that, when the data after spatio-temporal decomposition, the vehicle speeds at a certain intersection always remain in a relatively stable state, and those abnormal traffic series are easier to distinguish. The second component of our framework is a multi-head transformer model (CompFormer), which establishes connections between recent observations and historical traffic series stored in the data bank through a spatial attention matrix. Benefiting from the spatio-temporal decomposition of traffic series, spatial attention matrix can integrate all historical data with lower computing resources than spatio-temporal attention matrix to enhance the prediction ability of the model under extreme events. The architecture of our framework is depicted in and Figure~\ref{fig:compformer} and Figure~\ref{fig:framework}. The compensated features learned by CompFormer can be concatenated with the input embeddings, thus all the modules can be trained end-to-end.

Our modules can be flexibly integrated with existing traffic forecasting methods, such as DCRNN~\cite{dcrnn2017}, MTGNN~\cite{mtgnn2020}, GWN~\cite{graphwave2019}, GTS~\cite{gts2021}, DGCRN~\cite{dgcrn2021}) and STEP~\cite{step2022kdd}. We evaluate our proposed framework on two benchmark datasets, METR-LA and PEMS-BAY. The experimental results show that our method significantly improves prediction performance in extreme cases compared to six strong baselines, with an overall improvement of up to 28.2\%. Furthermore, we provide a detailed analysis of the effectiveness of our proposed CompFormer component, including ablation studies.

The main contributions of this paper are summarized as follows:
\begin{itemize}
    \item We identify extreme events as a critical challenge in multivariate traffic series forecasting and propose two evaluation metrics to measure the severity of these events.
    
    \item We develop a test-time compensated representation learning framework, consisting of a spatio-tempeoral decomposed data bank and a multi-head spatial transformer model (CompFormer), to address the issue of extreme events.

    \item Our framework can be flexibly integrated with existing forecasting methods, and we demonstrate its effectiveness on the METR-LA and PEMS-BAY benchmarks. We conduct extensive experiments and analyses to validate the performance of our method, showing significant improvements over six strong baselines, especially in extreme cases.

\end{itemize}

Overall, this paper provides a novel perspective on the problem of traffic forecasting and proposes an effective method for addressing extreme events, which is an under-explored area in the literature.

\section{Related Works}
\label{sec:related-work}
In this section, we first review the state-of-the-art deep learning models for traffic series forecasting. Next, we introduce the extreme events in general time series prediction scenarios. Finally, we describe how existing methods use the periodicity information to enhance the predictive capabilities.

\subsection{Traffic Speed Forecasting}
In recent years, Graph Neural Networks (GNNs) have become a frontier in traffic forecasting research and shown state-of-the-art performance due to their strong capabilities in modeling spatial dependencies from graphical datasets~\cite{dcrnn2017,graphwave2019,mtgnn2020,agcrn2020,stemm-gnn2020nips,dgcrn2021,gts2021,dmstgcn2021kdd,stgode2021kdd}. Most existing methods focus on modelling spatial dependencies by developing effective techniques to learn from the graph constructed from the geometric relations among the roads.
For example, with graphs as the input, the diffusion convolutional recurrent neural network (DCRNN)~\cite{dcrnn2017} models the spatial dependency of traffic series with bidirectional random walks on a directed graph as a diffusion process. Graph WaveNet (GWN)~\cite{graphwave2019} combines the self-adaptive adjacency matrix and dilated causal convolution to learn the spatial and temporal information respectively. 
However, the above models assume fixed spatial dependencies among roads~\cite{st-grat2020cikm}. Wu et al.~\cite{mtgnn2020} propose a general GNN-based framework for capturing the spatial dependencies without well-defined graph structure.

Benefiting from the ability of capturing global sequential dependency and parallelization, attention mechanism has become a popular technique in sequential dependency modeling, including self-attention~\cite{gman2020aaai,st-grat2020cikm,sttn2020,potionattn-2020www} and other variants~\cite{dual2017aaai,geoman2018ijcai,periodicity-revisit2019aaai}. Recent traffic forecasting models utilized multi-head attention~\cite{attention2017nips} to model spatial and temporal dependencies~\cite{gman2020aaai,st-grat2020cikm,sttn2020,potionattn-2020www,dual2017aaai,geoman2018ijcai,periodicity-revisit2019aaai}. Attention modules can directly access past information in long input sequences, but they cannot enlarge the predictor horizon to the whole historical data. This is because the length of the input need to be continuously increase, which results in intractable computational and memory cost in attention. Therefore, to enlarge the predictor  horizon of predictor without increasing the computing resource, we propose a spatio-temporal decomposition method in this paper that makes transformer only pay attention to spatial dimension, greatly reducing the complexity of compensated representation learning.

\subsection{Extreme Events}
Previous traffic forecasting methods overlook the extreme events that feature abrupt speed change, irregular and rare occurrences, resulting in poor performance when applying them for speed prediction. In other time series prediction tasks, most of the existing methods consider extreme events as a rare event prediction problem~\cite{extreme-1998icml,extreme2019kdd}. Ding et al.~\cite{extreme2019kdd} addressed the extreme events by formulating it as a data imbalance problem and resolving them by sample re-weighting. However, the extreme condition problem is more severe in traffic data due to many observations containing many zero-valued speeds and large entropy (Figure~\ref{fig:extreme}), which makes it difficult to give a proper definition for these extreme events to identify them during training because of the spatio-temporal complexity of traffic series. Therefore, we cannot address this problem by simply re-weighting as the above methods. In this paper, we construct a spatial transformer model to automatically extract compensated features from the whole historical data for the extreme events when needed.

\subsection{Periodicity}
Periodicity has been widely used in many time series prediction tasks. Considering the obvious periodicity of traffic data, Guo et al.~\cite{periodicity-attn2019aaai} proposed three independent spatial-temporal attention modules to respectively model three temporal properties of traffic flows, i.e., recent, daily-periodic and weekly-periodic dependencies. ST-ResNet~\cite{periodicity-stresnet2017aaai} is proposed for crowd flows prediction, in which three residual networks are used to model the temporal closeness, period, and trend properties of crowd traffic. Liang et al.~\cite{periodicity-urbanfm2019kdd} regarded the periodicities as a external factor and Yao et al.~\cite{periodicity-revisit2019aaai} designed a periodically shifted attention mechanism to handle long-term periodic temporal shifting.
However, their prediction ability is limited because of their limited horizon on the historical data.
Therefore, in order to make full use of whole historical data with limited computing resource, we propose the spatio-temporal decomposition method according to the periodicity characteristic.

\section{Extreme Events}
In this section, we first describe the mathematical expression of multivariate time series prediction in the context of traffic speed forecasting. Next, two typical types of extreme events that pose challenges for existing forecasting methods are introduced. Finally, we introduce two evaluation metrics to measure the extreme events.

\subsection{Preliminaries}
\label{sec:preliminaries}

Consider a multivariate traffic time series with $N$ correlated variables, denoted as $\mathbf{\mathcal{X}} =\{\mathbf{x}_{:,1}, x_{:,2}, \ldots, \mathbf{x}_{:,t}, \ldots\}$. Here, each component  $\mathbf{x}_{:,t} = (\mathbf{x}_{1,t}, \mathbf{x}_{2,t}, \ldots, \mathbf{x}_{N,t})^\top \in \mathbb{R}^{C\times D}$ represents the recording of $C$ sensors as time step $t$, where $D$ is the representation dimension of these sources. The goal is to predict the future values of this series based on the historical observations. Existing methods always formulate the problem as to find a function $\mathcal{F}(\cdot;\theta)$ to forecast the next $\tau$ steps based on the historical data of the past $L$ steps:
{\small
\begin{align}
    (\hat{\mathbf{x}}_{:,t+1}, \hat{\mathbf{x}}_{:,t+2}, \ldots, \hat{\mathbf{x}}_{:,t+\tau}) = \mathcal{F}(\mathbf{x}_{:,t-L+1}, \ldots, \mathbf{x}_{:,t-1}, \mathbf{x}_{:,t}; \theta), \label{eqn:general-prediction}
\end{align}
}
where $\theta$ is the parameters of the forecasting model.

We denote the input sequence at time $t$ as $\mathbf{X}_{t} =(\mathbf{x}_{:,t-L+1}, \ldots, \mathbf{x}_{:,t-1}, \mathbf{x}_{:,t})^\top \in \mathbb{R}^{L\times C \times D}$, where $L$ is the input length. In this paper, we take the traffic speed forecasting problem as an example and assume $D=2$, with one dimension recording the vehicle speed and the other dimension recording the global time stamp.

\begin{figure*}[t] 
    \centering
    \vspace{5pt}
    \includegraphics[scale=0.325]{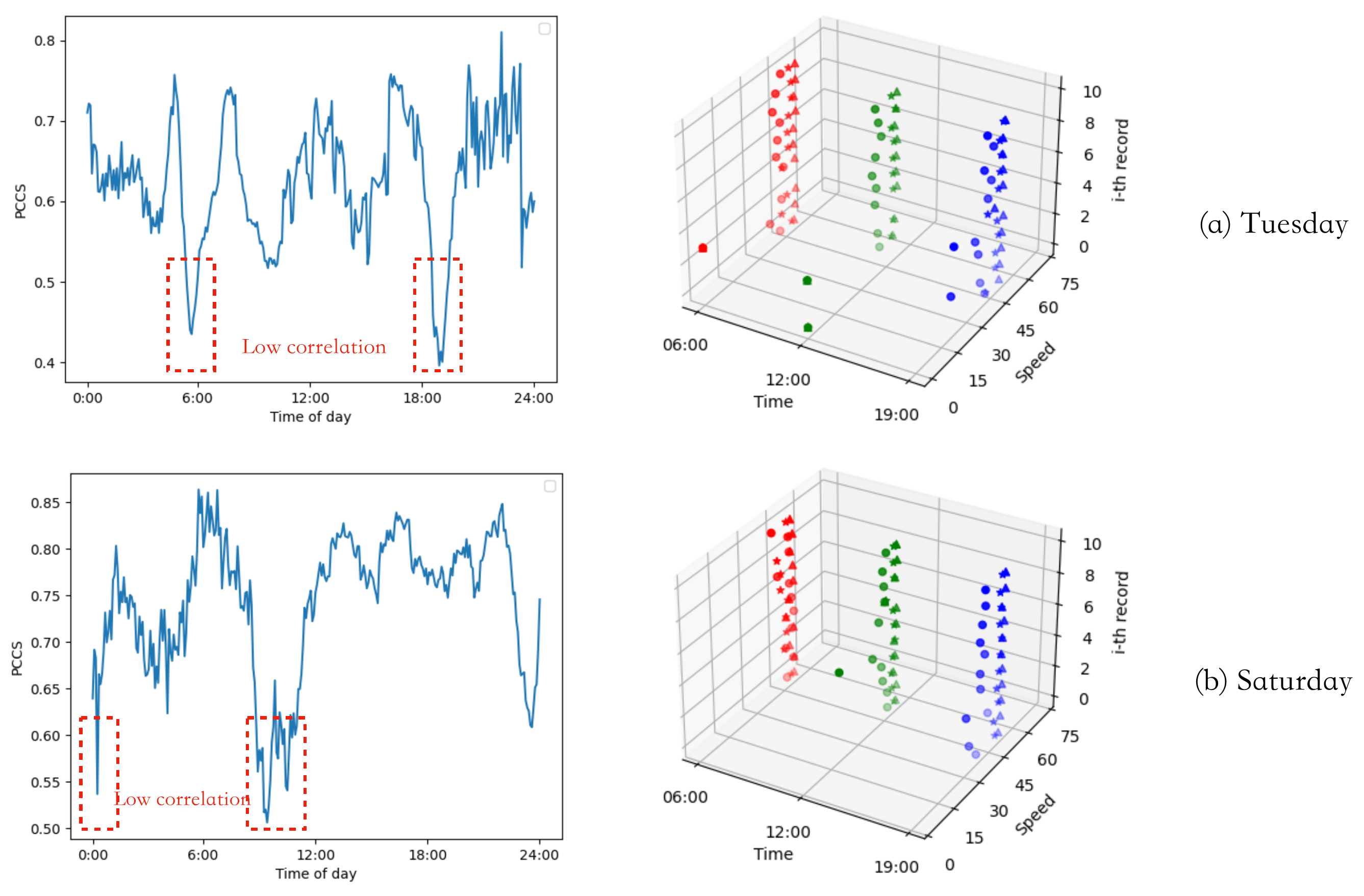}
    \caption{The visualization of Pearson product-moment correlation coefficients (PCCS) and historical traffic series after spatio-temporal decomposition on Tuesday and Saturday. 1) The first column of figures show that the traffic time series on weekdays and weekends presents different morning and evening peaks. 2) The second column of figures show that it is easy to identify the outliers (extreme events) of the spatio-temporal-decomposed traffic series at a certain intersection, which will reduce the difficulty of model prediction.}
    \label{fig:vis_pccs_3d}
\end{figure*}

\subsection{Definition of Extreme Events}
\label{sec:vis-series}
In this section, we provide visualization of traffic series to present the extreme events. As shown in Figure~\ref{fig:extreme}, we define extreme events by the distribution of observed and predicted values, as well as current road conditions. 
\begin{itemize}
    \item \textbf{Normal condition.} In non-morning and evening rush hours and without traffic jams, vehicle speeds at each intersection are stable at around 60 km/h, and the distributions of observations and predictions are similar, which enables the model to accurately predict the speed.

    \item \textbf{Extreme event [1].} The first extreme case shows that the linear correlation between observations and predictions is very low, such as having no correlation at all, which makes it impossible for the model to obtain accurate predictions only by relying on recent observations. 

    \item \textbf{Extreme event [2].} In the second extreme case, traffic jams lead to very large differences in vehicle speeds at various intersections (0 km/h - 70 km/h), and this infrequent event will also make it more difficult for the model to predict.
\end{itemize}

\subsection{Evaluation Metrics of Extreme Events}
Based on the definition of extreme events, we introduce two different evaluation metrics to measure the severity of extreme events. We also qualitatively analyze the relationship between evaluation metrics and prediction errors through visualization.

We denote $Z_t$ to be the number of zero speed in $\mathbf{X}_t$, which can be calculated as:
\begin{equation}
    Z_t = \sum_{l=1}^L \sum_{n=1}^C z_l^n,
\end{equation}
where $z_t^c=1$ if the speed of the $c$-th sensor at time $t-l+1$ is 0, i.e., $X_t(l, c, 0) = 0$, otherwise $z_l^n=0$. A larger $Z_t$ may be caused by a large area of traffic jam due to a traffic accident or the absence of vehicles in this time step, and this infrequent condition makes the speed prediction difficult.

The extreme event is also measured by the entropy of the speed values of the sensors in $\mathbf{X}_t$, which can be computed by:
\begin{equation}
    P_t = \frac{1}{L}\sum_{l=1}^L\sum_{c=1}^C - (\hat{\mathbf{X}}_t(l,c,0)+\epsilon) \text{log} (\hat{\mathbf{X}}_t(l,c,0)+\epsilon),
\end{equation}
where $\hat{\mathbf{X}}_t$ is the normalized probability distribution, and $\epsilon > 0$ is a small positive value used to address the zero-valued speed for numerical stability. A larger $P_t$ indicates a more diverse speed distribution, which implies a higher level of traffic congestion and more difficulty in predicting speeds.

As shown in Figure~\ref{fig:extreme} (a), the distribution curves of the two evaluation metrics exhibit long-tailed distributions, which is a good indication that extreme events are infrequent but repetitive. About 0.4\% of the recorded values in the multidimensional traffic speed series are all zeros. The empirical results shown in Figure~\ref{fig:extreme} (b) demonstrate that existing methods perform poor in these events. In the 9th, 84th and 103th predictions, the larger $Z_t$ and $P_t$ indicate that the extreme events are more serious, and the prediction error of the model is correspondingly larger. Therefore, we would like to point out that predicting traffic speed in the above two infrequent but repetitive extreme events is important and valuable in smart transportation systems, especially for driving route planning and vehicle dispatching. However, traffic speed forecasting that rely solely on recent observations is unreliable, especially under extreme events.

\section{Methodology}
\label{sec:method}
In this section, we first visualize the traffic time series before and after spatio-temporal decomposition to analyze why the following proposed periodic data bank is necessary. Next, we introduce the construction of periodic data bank. Finally, our proposed CompFormer is presented to learn compensated representation.

\subsection{Spatio-temporal Decomposition}
The existence of extreme events renders the predictions unreliable by relying solely on recent observations. Therefore, it is worth exploring how to use all historical data to participate in the testing phase to address the aforementioned shortcomings. Our analysis indicates that using more historical data as observations (such as all training data, one week's data and one day's data) for prediction would result in the inclusion of significant redundant information, which will greatly increase the difficulty of model prediction.

We visualize and analyze the correlation of traffic time series among different adjacent time stamps by calculating their correlation coefficients. In order to analyze the linear correlation of $\mathbf{X}_t$, we compute their pearson product-moment correlation coefficient (PPMCC) by:
\begin{equation}
\begin{split}
    \rho (\mathbf{X}_t(l-1, :, 0), \mathbf{X}_t(l, :, 0)) = \\
    \frac{\sum_{i=1}^C (x_{ti}^{l-1}-\Bar{x}_t^{l-1}) (x_{ti}^{l}-\Bar{x}_t^{l})}{\sqrt{\sum_{i=1}^C (x_{ti}^{l-1}-\Bar{x}_t^{l-1})^2} \sqrt{\sum_{i=1}^C (x_{ti}^{l}-\Bar{x}_t^{l})^2}}
\end{split}
\end{equation}
where $x_{t}^{l} \in \mathbb{R}^{C}$ is speed values of all intersections at time stamp $t$. Thus, the value of $\rho$ can vary between $-1$ and $1$. The smaller absolute value $|\rho|$ means that the $x_{t}^{l-1}$ and $x_{t}^{l}$ at the adjacent time are less linearly correlated, which will increase the predictive difficulty of the model.

\begin{figure*}[t] 
    \centering
    \vspace{5pt}
        \includegraphics[scale=0.625]{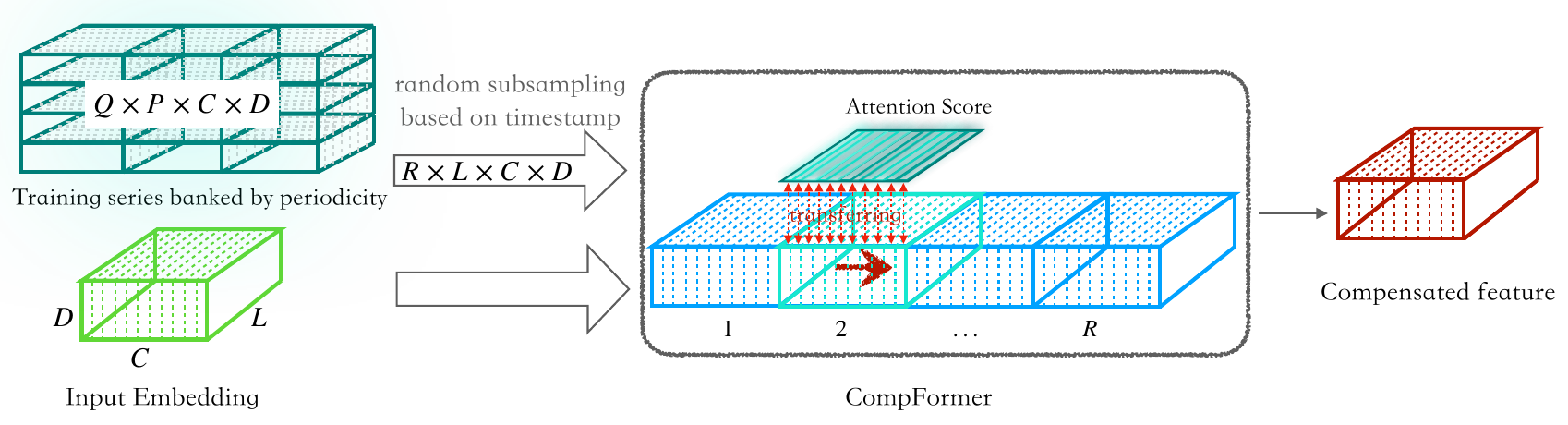}
    \caption{The framework of compensation representation learning process. 1) The input embedding is set as query (Q), and the selected $R$ historical series are presented as key (K) and value (V). 2) The size of compensated feature learned by CompFormer is equal to input (i.e., query).}
    \label{fig:framework}
\end{figure*}

From the first column of Figure~\ref{fig:vis_pccs_3d}, we can see that the correlation coefficients of the multi-dimensional traffic time series of five minutes before and after show obvious morning and evening peaks on weekdays. The low correlation coefficient values mean that long-term predictions using only short-term historical data are unreliable.
Another conclusion is that traffic conditions at various times on weekends are different from those on weekdays. If we simply enlarge the input of the model to the historical data of the past week or past day, this requires a lot of computing resources to make the model find informative data from redundant information~\cite{step2022kdd}. Therefore, how to use all historical data for prediction under limited computing resources is an urgent problem to be solved.

As shown in the second column of Figure~\ref{fig:vis_pccs_3d}, the outliers in highly complicated traffic series are easy to identify after spatio-temporal decomposition. For example, at 06:00, the speed values recorded at the intersection are within a stable range, $\approx 60km/h$. In addition, an intersection contains only about ten recorded values at a particular time in the entire METR-LA dataset, thus allowing for accurate predictions to be made with minimal computational resources. The next question is how to use all the historical data after spatio-temporal decomposition. Therefore, we propose the following periodic data bank and the spatial transformer model.

\subsection{Periodic Data Bank}
In this section, we propose to separate all the historical traffic series according to the periodicity characteristic to construct a periodic data bank, which includes the following data preprocessing and bank construction.

\textbf{Data preprocessing.} We replace all zero-valued speeds in the training data by looking up the historical data. The zero-valued speeds have two different types. The first is all the speed values of input traffic series equal to zero, thus we replace them with the data at the same periodic time. The second is only a small part of input sequence has zero speed, we replace them with mean value of the remain non-zero values.

\textbf{Bank construction.}
In constructing the periodic data bank, we consider the various periodic granularities of traffic series, such as $P=2016$ is the periodicity combined from minute of hour (12), hour of day (24) and day of week (7). By dong this, the numbers of records in one periodicity is $Q$. 
We partitions and stores all the traffic series into a matrix according to their periodicity.
\begin{equation}
    \mathbf{\mathcal{M}} = [\mathbf{M}_{1}, \mathbf{M}_{2}, ..., \mathbf{M}_{P}]^\top \in \mathbb{R}^{P \times Q \times C}
\end{equation}
where $\mathbf{M}_{p} \in \mathbb{R}^{Q \times C}$, each column stores all the recorded data of a certain periodicity. For example, as shown in Figure~\ref{fig:vis_pccs_3d}, at 6:00 am on Tuesday, a total of $Q=10$ pieces of data were recorded at a certain intersection, and the METR-LA dataset has $C=207$ intersections, so $\mathbf{M}_{\text{Tues, 06:00}} \in \mathbb{R}^{10 \times 207}$.


One of the benefits of this spatio-temporal decomposition is that each $\mathbf{\mathbf{M}}_p$ slice functions as a representative of all the training data collected for a given periodicity, thereby reducing the computational complexity of the subsequent transformer model designed to learn compensated representation.

\begin{figure*}[t]
\vspace{5pt}
    \centering
        \includegraphics[scale=0.35]{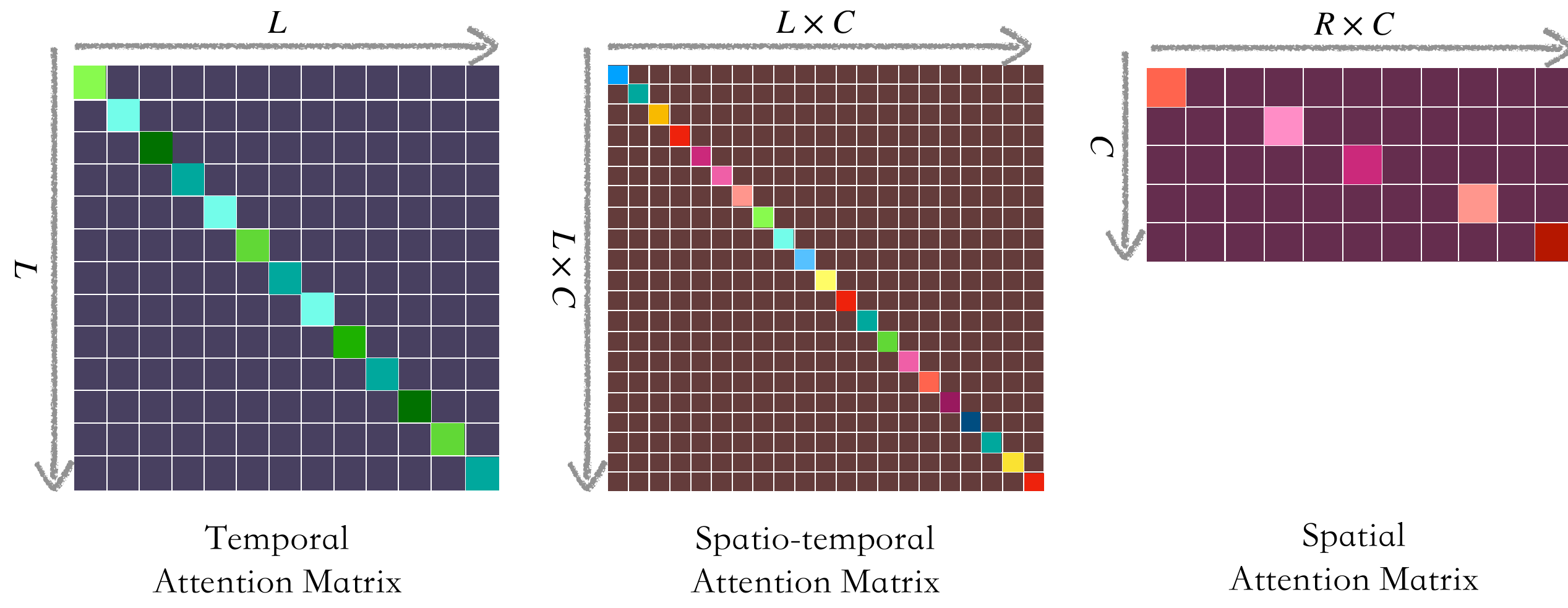}
    \caption{The visualization of three different types of attention matrices.}
    \label{fig:attn_matrix}
\end{figure*}

\begin{algorithm}[t]
\caption{Test-Time Compensated Representation Learning Framework}

\flushleft{\textbf{Required Model:} Spatial transformer model $\mathcal{A}_\theta$, forecasting model $\mathcal{F}_\theta$}
\flushleft{\textbf{Required Input:} Mini batch input $\mathbf{X} \in \mathbb{R}^{B \times L \times C \times D}$} 

\textbf{Required Periodic Data Bank:} $\mathcal{M} \in \mathbb{R}^{P \times Q \times C \times D}$ \\

    \begin{algorithmic} [1]
    
    \FOR{$i$=$1$ to $L$}

    \STATE{Current input $\mathbf{X}_i^t \in \mathbb{R}^{B \times C \times 1}$} 

    \STATE{Extract series from $\mathcal{M}$: $M_{i}^{t+1} \in \mathbb{R}^{B \times (R \times C) \times 1}$}
    \STATE{Set Query $\mathbf{X}_i^t$, Key $M_{i}^{t+1}$, Value $M_{i}^{t+1}$}
    \STATE{Attention output $O_i^{t+1} \in \mathbb{R}^{B \times C \times D} = A_\theta(\mathbf{X}_i, M_{i}^{t+1}, M_{i}^{t+1})$}
    \STATE{Assign $F_{o}^{t+1}[i]=O_i^{t+1}$}
    \ENDFOR
    
    \STATE{New input $\mathbf{X} \in \mathbb{R}^{B \times L \times C \times (D+D)}$ = \textit{concat}($\mathbf{X}, \mathbf{F}_{o}^{t+1}$)}
    
    \STATE{Prediction $Y = f_\theta(\mathbf{X})$}
    \STATE{Compute MAE loss}
    \STATE{Update $\mathcal{A}_\theta$ and $\mathcal{F}_\theta$ by back-propagation}
    
    \end{algorithmic}
\label{alg:algorithm}
\end{algorithm}

\begin{figure*}[t]
\centering
\vspace{-12pt}
\subfigure[Results from GWN~\cite{graphwave2019}]{\includegraphics[scale=0.55]{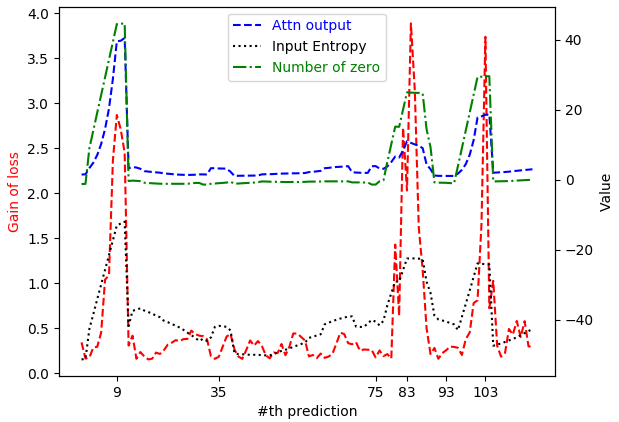}} \hspace{5pt}
\subfigure[Results from DGCRN~\cite{dgcrn2021}]{ \includegraphics[scale=0.55]{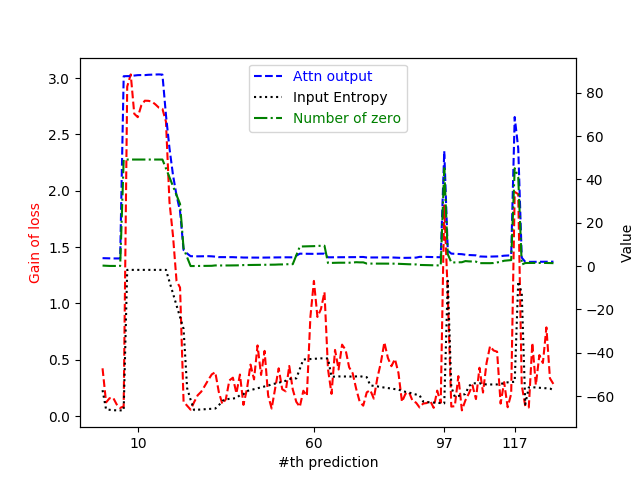}}
\caption{The visualization of the relationship among the loss gap with and without our method. We can see that in the events with higher degree of extremeness, the improvement achieved by our module becomes more significant, i.e, the loss gap becomes larger.}
\label{fig:loss-gap}
\end{figure*}

\begin{figure*}[t]
\centering
{\includegraphics[scale=0.5525]{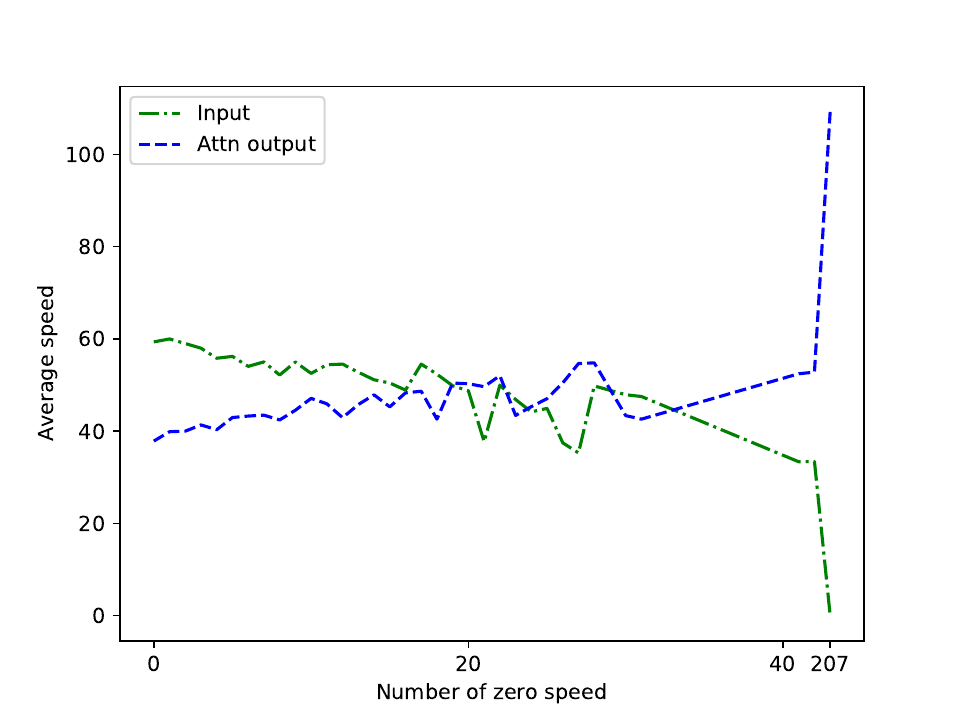}} \hspace{0pt}
{\includegraphics[scale=0.5525]{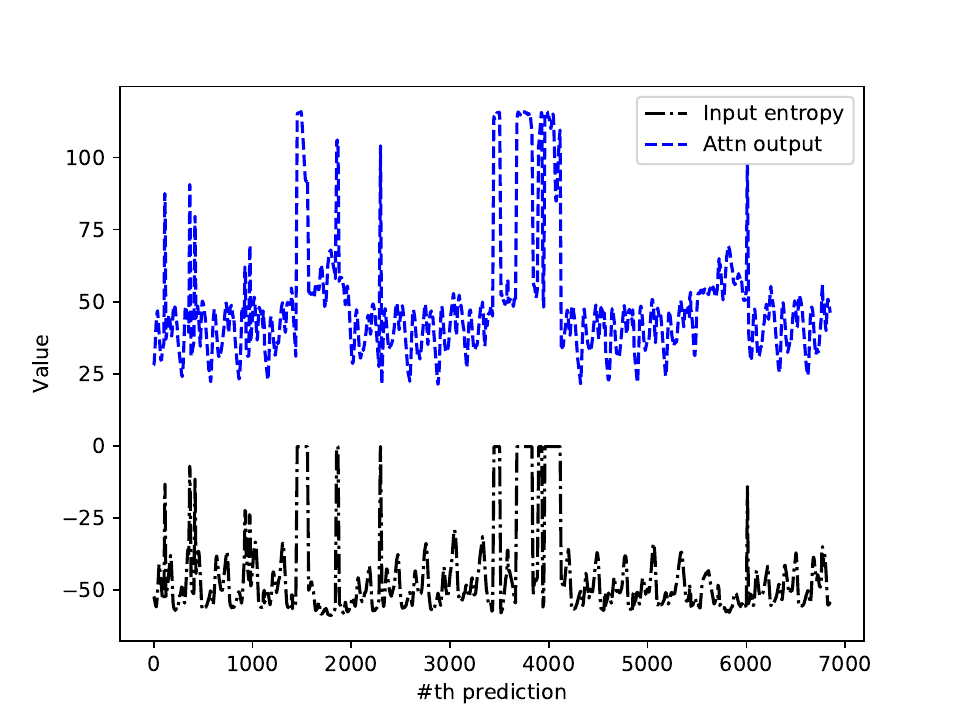}}
\caption{The visualization of relationship between the effect of our spatial attention module and the extremeness. The results show that the magnitude of the attention output rapidly increases when the number of zero-valued speed or the entropy of the input becomes larger.}
\label{fig:input_attn_zero_entropy}
\end{figure*}

\begin{table}[t]
    \vspace{8pt}
    \centering
    \caption{Complexity comparison with different attention mechanism. Here $L$ denotes the historical sequence length, $C$ is the spatial dimension, $R$ is number of hitorical series extracted from the periodic data bank, where $R \ll L$. We assume the feature dimension is $1$.}
    \scalebox{1.08}{
    \begin{tabular}{lcc}
    \toprule
    Layer Type    & Time    & Space  \\
        
    \hline
    Temporal Attention           & $\mathcal{O}(L^2)$   & $\mathcal{O}(L^2 + L)$  \\
        
    Spatio-temporal Attention    & $\mathcal{O}(L^2C^2)$   & $\mathcal{O}(L^2C^2 + LC)$  \\

    Spatial Attention            & $\mathcal{O}(\mathbf{RC^2})$   & $\mathcal{O}(\mathbf{RC^2 + RC + C})$  \\
    \bottomrule
\end{tabular}}
\label{tab:compare-attention}
\end{table}

\subsection{Spatial Transformer}
\label{sec:4.2}
Now we turn to present our multi-head spatial transformer module that can extract the compensated features from the above periodic data bank for extreme traffic prediction. After spatio-temporal decomposition, the proposed CompFormer only needs to automatically identify the anomalous patterns of $\mathbf{X}_{t} \in \mathbb{R}^{L\times C \times D}$ in $C$ dimension and transfer robust knowledge from periodic data bank based on the time stamp $t$.

To be precise, our CompFormer module is developed as follows. As shown in Figure~\ref{fig:framework}, the observations $\mathbf{X}_{t}(:, :, 0) \in \mathbb{R}^{L\times C \times 1}$ with extremeness are transformed to embeddings by fully connected layers, and the historical embeddings $\hat{\mathbf{X}}_{t+1} \in \mathbb{R}^{L \times RC \times D}$ are random sampled from periodic data bank based on time stamp $t$, where hyper-parameter $R$ is the number of sampled series. $\hat{\mathbf{X}}_{t+1}$ represents the future historical data, because the temporal dynamics can cause the road conditions to be very different in the five minutes before and after, especially in the morning and evening peaks.
In addition, we would like to point out that due to our well organized periodic data bank, such simple random sampling can achieve good performance, which is verified in our experiments.

In our proposed CompFormer, we take the short observation embeddings $\mathbf{X}_{t} \in \mathbb{R}^{L\times C \times D}$ and sampled historical series $\hat{\mathbf{X}}_{t+1} \in \mathbb{R}^{L \times RC \times D}$ as query ($Q$) and key ($K$), value ($V$), respectively. 
The computation of CompFormer takes the form of
\begin{equation}
\begin{split}
    a_{ij} &= \text{Softmax}\left(\frac{\text{exp}(s_{ij})}{\sum_{k=1}^{RC} \text{exp}(s_{ik})} \right) \\
    s_{ij} &= \mathcal{A}_\theta\left(\mathbf{X}_{t} (L, i, D), \hat{\mathbf{X}}_{t+1}(L, j, D) \right),
\end{split}
\end{equation}
where $\mathcal{A}_\theta$ is the transformer model, $i=1,2, \dots C, j=1,2, \dots RC$. $a \in \mathbb{R}^{C \times RC}$ is the \textit{spatial attention matrix}. The attention score $a_{ij}$ represents the relationship between the input and the historical data in spatial dimension, i.e., its value shows the importance of the corresponding sampled series from periodic data bank, and it guides the interaction of information between them to transfer robust representation from $\hat{\mathbf{X}}_{t+1}$. 
Next,
\begin{equation}
    O_{r} = \sum_{r=1}^{R} a_{r} \times \hat{\mathbf{X}}_{t+1}(r),
\end{equation}
after finishing the above computation process, the CompFormer output $O \in \mathbb{R}^{L \times C \times D}$ (compensated features) will be concatenated on the input $\mathbf{X}_t^{L \times C \times D}$ to be put into the forecasting model, as illustrated in Figure~\ref{fig:framework}. The detailed steps of our method are given in Algorithm~\ref{alg:algorithm}.

\textbf{Complexity comparison.} As illustrated in Figure~\ref{fig:attn_matrix}, the traditional temporal attention calculates the relative weight between $L$ sequences, and its time complexity is $L^2$. Multivariate traffic series need to learn both spatial dependence and temporal dynamics, so the time complexity of spatio-temporal
attention is $L^2C^2$. However, due to the very high computational complexity and memory consumption, we cannot simultaneously learn the importance among very long historical sequences. Due to the spatio-temporal decomposed data bank we constructed, the time complexity of spatial attention is $RC^2$, and it no longer dependents on the length of input sequence. In our experiments, the value of $R$ is 5.

\begin{table*}[t]
    \vspace{10pt}
    \centering
    \caption{Statistics of datasets.}
    \scalebox{1.1}{
    \begin{tabular}{cccccc}
    \toprule
    Datasets & Samples & Nodes & Sample Rate & Input Length & Output Length \\
        
    \hline
    METR-LA  & 34272 & 207 & 5min & 12 & 12 \\
        
    PEMS-BAY & 52116 & 325 & 5min & 12 & 12 \\
    \bottomrule
\end{tabular}}
\label{tab:dataset}
\end{table*}

\begin{table*}[t]
\vspace{8pt}
\centering
\caption{The results on METR-LA and PEMS-BAY datasets. We integrate our method with the latest strong baselines DCRNN~\cite{dcrnn2017}, GWN~\cite{graphwave2019}, MTGNN~\cite{mtgnn2020}, GTS~\cite{gts2021}, DGCRN~\cite{dgcrn2021} and STEP~\cite{step2022kdd}.
}
\scalebox{1.1}{   
\begin{tabular}{ll|c|c|c}  
    \toprule
    \multirow{3}*{Methods}  & \multirow{3}*{Venue}  & \multicolumn{3}{c}{\textcolor{red}{METR-LA}}  \\

    \cline{3-5}
    &  &  \multicolumn{1}{c}{15 min}  &  \multicolumn{1}{c}{30 min}  &  \multicolumn{1}{c}{60 min} \\
		
    \cline{3-5}
    &  & MAE / RMSE / MAPE  & MAE / RMSE / MAPE  & MAE / RMSE / MAPE  \\

        \hline
        FC-LSTM  & -   & 3.44 / 6.30 / 9.60   & 3.77 / 7.23 / 10.90   & 4.37 / 8.69 / 13.20 \\
        
        WaveNet~\cite{wavenet2016}  & 2016 ICLR   & 2.99 / 5.89 / 8.04   & 3.59 / 7.28 / 10.25   & 4.45 / 8.93 / 13.62 \\
        
        STGCN~\cite{stgcn2017} & 2017 IJCAI & 2.88 / 5.74 / 7.62    & 3.47 / 7.24 / 9.570    & 4.59 / 9.40 / 12.70 \\

        ST-MetaNet~\cite{st-metanet2019} & 2019 KDD & 2.69 / 5.17 / 6.91 & 3.10 / 6.28 / 8.57 & 3.59 / 7.52 / 10.63 \\

        LDS~\cite{lds2019}   & 2019 ICML    & 2.75 / 5.35 / 7.10    & 3.14 / 6.45 / 8.60     & 3.63 / 7.67 / 10.34 \\
        
        AGCRN~\cite{agcrn2020} & 2020 NeuIPS  & 2.87 / 5.58 / 7.70    & 3.23 / 6.58 / 9.000    & 3.62 / 7.51 / 10.38 \\
        
        GMAN~\cite{gman2020aaai}  & 2020 AAAI    & 2.80 / 5.55 / 7.41    & 3.12 / 6.49 / 8.730    & 3.44 / 7.35 / 10.07 \\

        \cline{1-5}
        DCRNN~\cite{dcrnn2017}  & 2018 ICLR   & 2.77 / 5.38 / 7.30    & 3.15 / 6.45 / 8.80     & 3.60 / 7.59 / 10.50   \\

        DCRNN \textit{w / Ours (CompFormer)} & -   &  \textbf{2.56} / \textbf{5.15} / \textbf{6.60}   & \textbf{2.98} / \textbf{6.34} / \textbf{8.01}    & \textbf{3.53} / \textbf{7.34} / \textbf{9.610}   \\

        \cdashline{3-5}[1pt/1pt]
        GWN~\cite{graphwave2019}  & 2019 IJCAI  & 2.71 / 5.16 / 7.11    & 3.11 / 6.24 / 8.54    & 3.56 / 7.31 / 10.10   \\
        
        GWN \textit{w / Ours (CompFormer)}  & - & \textbf{2.67} / \textbf{5.09} / \textbf{6.83}    & \textbf{3.03} / \textbf{6.07} / \textbf{8.19}    & \textbf{3.43} / \textbf{7.06} / \textbf{9.680} \\

        \cdashline{3-5}[1pt/1pt]
        MTGNN~\cite{mtgnn2020}  & 2020 KDD   & 2.69 / 5.18 / 6.86    & 3.05 / 6.17 / 8.19     & 3.49 / 7.23 / 9.87   \\

        MTGNN \textit{w / Ours (CompFormer)} & -   & \textbf{2.65} / \textbf{5.11} / \textbf{6.78}   & \textbf{3.00} / \textbf{6.07} / \textbf{8.11}    & \textbf{3.40} / \textbf{7.09} / \textbf{9.65}   \\

        \cdashline{3-5}[1pt/1pt]
        GTS~\cite{gts2021}$^*$   & 2021 ICLR   & 2.92 / 5.91 / 7.70   & 3.51 / 7.29 / 10.0    & 4.28 / 8.87 / 13.10   \\
        
        GTS \textit{w / Ours (CompFormer)} & -   & \textbf{2.62} / \textbf{5.18} / \textbf{6.70}   & \textbf{3.00} / \textbf{6.22} / \textbf{7.90}    & \textbf{3.39} / \textbf{7.26} / \textbf{9.400}   \\

        \cdashline{3-5}[1pt/1pt]
        DGCRN~\cite{dgcrn2021}  & 2021 TKDE   & 2.64 / 5.07 / 6.68    & 3.02 / 6.13 / 8.05    & 3.49 / 7.33 / 9.72   \\
        
        DGCRN \textit{w / Ours (CompFormer)} & -   & \textbf{2.59} / \textbf{5.00} / \textbf{6.58}    & \textbf{2.96} / \textbf{6.05} / \textbf{7.89}    & \textbf{3.38} / \textbf{7.12} / \textbf{9.51}   \\

        \cdashline{3-5}[1pt/1pt]
        STEP~\cite{step2022kdd}  & 2022 KDD & 2.61 / 4.98 / 6.60    & 2.96 / 5.97 / 7.96    & 3.37 / 6.99 / 9.61   \\
        
        STEP \textit{w / Ours (CompFormer)}  & -   & \textbf{2.58} / \textbf{4.94} / \textbf{6.47}   & \textbf{2.85} / \textbf{5.82} / \textbf{7.80}    & \textbf{3.27} / \textbf{6.76} / \textbf{9.46}   \\

    \hline
    \hline
    \multirow{3}*{Methods} & \multirow{3}*{Venue}   & \multicolumn{3}{c}{\textcolor{red}{PEMS-BAY}} \\

    \cline{3-5}
    & &  \multicolumn{1}{c}{15 min}  &  \multicolumn{1}{c}{30 min}  &  \multicolumn{1}{c}{60 min} \\
		
    \cline{3-5}
    & &  MAE / RMSE / MAPE  & MAE / RMSE / MAPE  & MAE / RMSE / MAPE  \\
        
    \hline   
        FC-LSTM  & -    & 2.05 / 4.19 / 4.80    & 2.20 / 4.55 / 5.20    & 2.37 / 4.96 / 5.70 \\
        
        WaveNet~\cite{wavenet2016}  & 2016 ICLR   & 1.39 / 3.01 / 2.91    & 1.83 / 4.21 / 4.16    & 2.35 / 5.43 / 5.87 \\
        
        STGCN~\cite{stgcn2017} & 2017 IJCAI   & 1.36 / 2.96 / 2.90    & 1.81 / 4.27 / 4.17    & 2.49 / 5.69 / 5.79 \\
        
        AGCRN~\cite{agcrn2020} & 2020 NeuIPS   & 1.37 / 2.87 / 2.94    & 1.69 / 3.85 / 3.87    & 1.96 / 4.54 / 4.64 \\
        
        GMAN~\cite{gman2020aaai}  & 2020 AAAI   & 1.34 / 2.91 / 2.86    & 1.63 / 3.76 / 3.68    & 1.86 / 4.32 / 4.37 \\

    \cline{1-5}
        DCRNN~\cite{dcrnn2017}  & 2018 ICLR  & 1.38  / 2.95 / 2.90    & 1.74 / 3.97  / 3.90    & 2.07  / 4.74 / 4.90   \\

        DCRNN \textit{w / Ours (CompFormer)}  & -   & \textbf{1.32}  / \textbf{2.78} / \textbf{2.76}   & \textbf{1.65} / \textbf{3.73}  / \textbf{3.64}    & \textbf{1.95}  / \textbf{4.49} / \textbf{4.55}   \\

    \cdashline{3-5}[1pt/1pt]
        GWN~\cite{graphwave2019}  & 2019 IJCAI   & 1.33  / 2.75 / 2.71    & 1.65 / 3.75  / 3.73    & 2.00  / 4.63 / 4.77   \\
        
        GWN \textit{w / Ours (CompFormer)}  & -   & \textbf{1.31} / \textbf{2.73} / \textbf{2.70}   & \textbf{1.60} / \textbf{3.63} / \textbf{3.62}    & \textbf{1.92} / \textbf{4.38} / \textbf{4.48}   \\

    \cdashline{3-5}[1pt/1pt]
        MTGNN~\cite{mtgnn2020}  & 2020 KDD   & 1.34  / 2.83 / 2.84   & 1.66 / 3.79 / 3.77  & 1.95  / 4.52 / 4.64   \\

        MTGNN \textit{w / Ours (CompFormer)}  & -   & \textbf{1.32}  / \textbf{2.78} / \textbf{2.76}   & \textbf{1.64} / \textbf{3.68}  / \textbf{3.65}    & \textbf{1.92}  / \textbf{4.38} / \textbf{4.44}   \\

    \cdashline{3-5}[1pt/1pt]
        GTS~\cite{gts2021}$^*$   & 2021 ICLR   & 1.34 / 2.83 / 2.86    & 1.67 / 3.78 / 3.83    & 1.95 / 4.45 / 4.67   \\
        
        GTS \textit{w / Ours (CompFormer)}  & -   & \textbf{1.31} / \textbf{2.80} / \textbf{2.79}   & \textbf{1.62} / \textbf{3.70} / \textbf{3.75}    & \textbf{1.87} / \textbf{4.33} / \textbf{4.52}   \\

    \cdashline{3-5}[1pt/1pt]
        DGCRN~\cite{dgcrn2021}  & 2021 TKDE   & 1.30 / 2.73 / 2.71    & 1.62 / 3.71 / 3.64    & 1.93 / 4.52 / 4.56   \\
        
        DGCRN \textit{w / Ours (CompFormer)}  & -   & \textbf{1.28} / \textbf{2.69} / \textbf{2.66}   & \textbf{1.57} / \textbf{3.64} / \textbf{3.55}    & \textbf{1.83} / \textbf{4.41} / \textbf{4.42}   \\

        \cdashline{3-5}[1pt/1pt]
        STEP~\cite{step2022kdd}  & 2022 KDD   & 1.26 / 2.73 / 2.59    & 1.55 / 3.58 / 3.43    & 1.79 / 4.20 / 4.18   \\
        
        STEP \textit{w / Ours (CompFormer)}  & -   & \textbf{1.24} / \textbf{2.70} / \textbf{2.48}   & \textbf{1.46} / \textbf{3.49} / \textbf{3.37}    & \textbf{1.69} / \textbf{4.08} / \textbf{4.05}   \\
        
    \bottomrule
\end{tabular}}
\label{tab:results}
\end{table*}

\section{Experiments}
\label{sec:exp}

\subsection{Datasets}
Our experiments are conducted on the following two most widely used real-world large-scale traffic datasets, METR-LA and PEMS-BAY, in which 70\% of data is used for training, 20\% are used for testing while the remaining 10\% for validation. METR-LA~\cite{dcrnn2017} is a public traffic speed dataset collected from loop detectors in the highway of Los Angeles containing 207 selected sensors and ranging from Mar 1st 2012 to Jun 30th 2012. The unit of speed is mile/h. PEMS-BAY~\cite{dcrnn2017} is another public traffic speed dataset collected by California Transportation Agencies (CalTrans) with much larger size. It contains 325 sensors in the Bay Area ranging from Jan 1st 2017 to May 31th 2017. The detailed information of the above two datasets is illustrated in Table~\ref{tab:dataset}.
 
Following the empirical settings of DCRNN~\cite{dcrnn2017}, we aim to predict the traffic speeds of next 15, 30 and 60 minutes based on the observations in the previous hour. As in both of these two datasets, the sensors record the speed value for every 5 minutes, thus the input length equals to $12$, while the output length equals to $3$, $6$ and $12$ in our experiments.

\subsection{Baselines and Configurations}
We verify the effectiveness of our proposed method in boosting existing methods by integrating it with the following six latest strong  baselines: \textbf{DCRNN}~\cite{dcrnn2017}, \textbf{MTGNN}~\cite{mtgnn2020}, \textbf{GWN}~\cite{graphwave2019}, \textbf{GTS}~\cite{gts2021}, \textbf{DGCRN}~\cite{dgcrn2021} and \textbf{STEP}~\cite{step2022kdd}. We also give the results of some other representative baselines to better evaluate the performance of our method.

All the experiments are implemented by Pytorch 1.7.0 on a virtual workstation with a 11G memory Nvidia GeForce RTX 2080Ti GPU. 
We train our model by using  Adam optimizer with gradient clip threshold equals to 5. 
The batch size is set to 64 on METR-LA and PEMS-BAY datasets. Early stopping is employed to avoid overfitting. We adopt the  MAE loss function defined in Section \ref{sec:metric} as the objective for training.

\begin{figure*}[t]
\centering
\vspace{8pt}
\begin{tabular}{c@{ }@{ }c@{ }@{ }c@{ }@{ }c}
    \begin{minipage}{0.5\linewidth}\centering
	\includegraphics[scale=0.52]{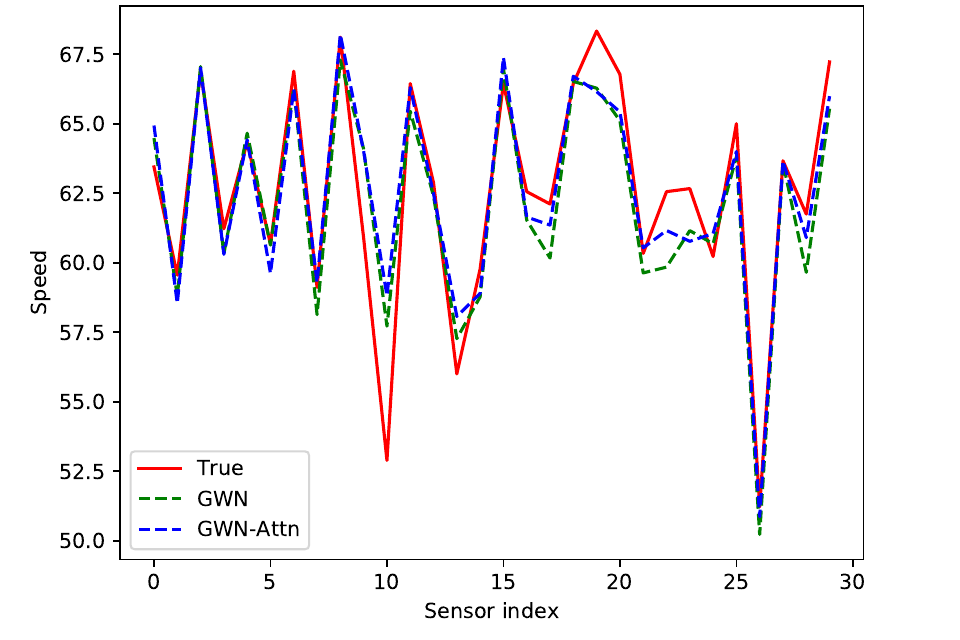} \\
        (a) Number of zero: 4, Entropy: 52.85
    \end{minipage}  &
    \begin{minipage}{0.5\linewidth}\centering
	\includegraphics[scale=0.52]{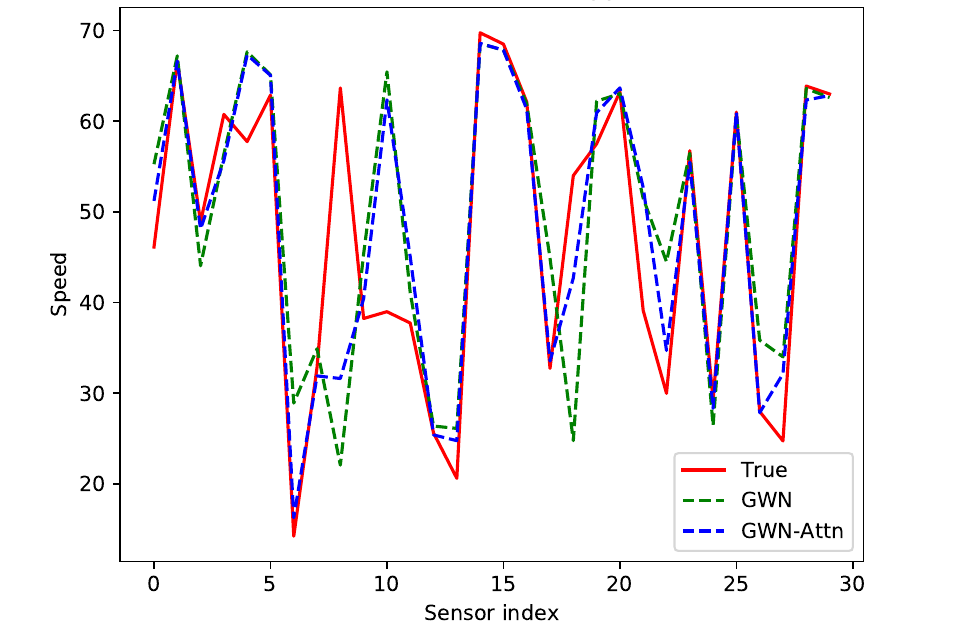} \\
        (b) Number of zero: 41, Entropy: 27.96
    \end{minipage} & \vspace{15pt} \\

    \begin{minipage}{0.5\linewidth}\centering
        \includegraphics[scale=0.52]{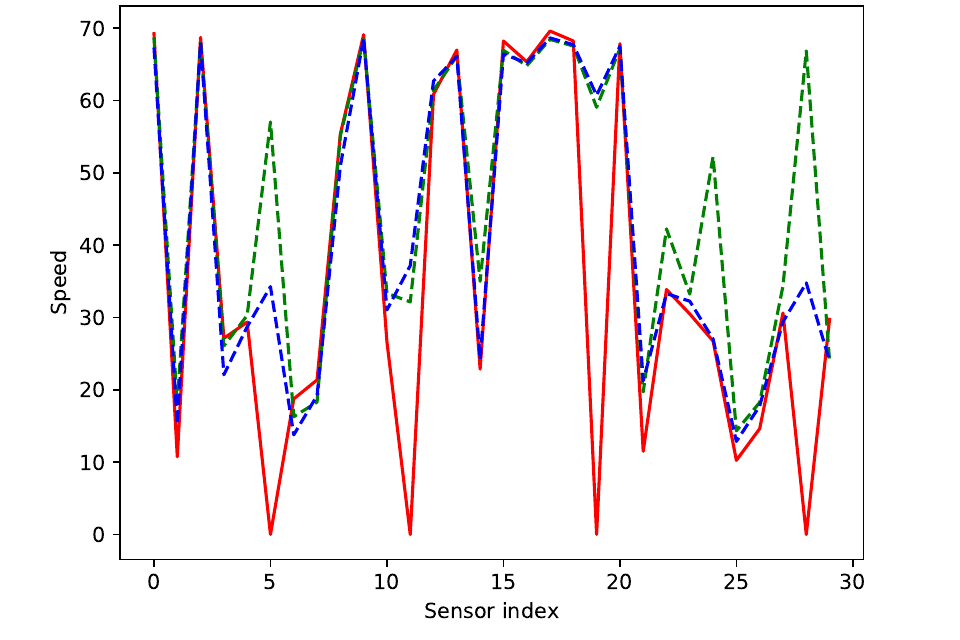} \\
        (c) Number of zero: 26, Entropy: 29.96
    \end{minipage} &
    \begin{minipage}{0.5\linewidth}\centering
        \includegraphics[scale=0.52]{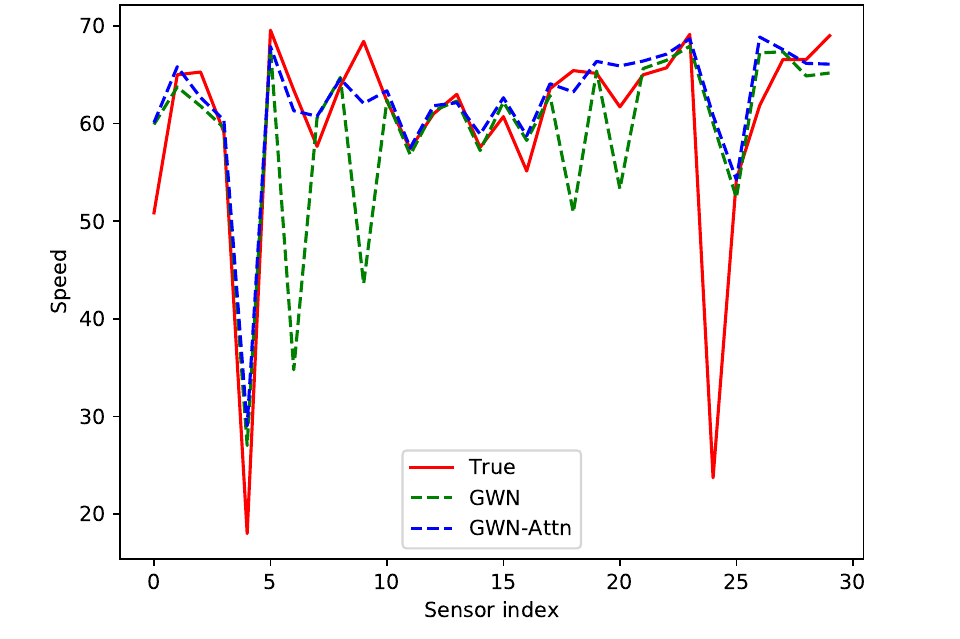} \\
        (d) Number of zero: 207, Entropy: 0.25
    \end{minipage} & \\
\end{tabular}
\vspace{1.5pt}
\caption{The visualization of performance of our method compared with GWN~\cite{graphwave2019} in the extreme events on METR-LA dataset. It shows that when the events become more extreme, the superiority of our method becomes more significant.}
\label{fig:vis-result-dgcrn}
\vspace{3.5pt}
\end{figure*}

\begin{table*}[t]
\vspace{5pt}
\centering
\caption{The experimental results of MTGNN~\cite{mtgnn2020} and DGCRN~\cite{dgcrn2021} on METR-LA dataset.}
\scalebox{1.05}{
	\begin{tabular}{llll|lll|lll}  
	    \toprule
		 \multirow{2}*{Methods}    &  \multicolumn{3}{c}{15 min}  &  \multicolumn{3}{c}{30 min}  &  \multicolumn{3}{c}{60 min} \\
		 
		\cline{2-10}
        & MAE & RMSE & MAPE  & MAE & RMSE & MAPE & MAE & RMSE & MAPE \\
        
        
        \hline
        MTGNN \textit{w / add}  & 2.70 & 5.22 & 7.07\%   & 3.04 & 6.19 & 8.37\%  & 3.43 & 7.18 & 9.86\%   \\
        
        MTGNN \textit{w / cat}  & \textbf{2.65} & \textbf{5.11} & \textbf{6.78\%}   & \textbf{3.00} & \textbf{6.07} & \textbf{8.11\%}  & \textbf{3.40} & \textbf{7.09} & \textbf{9.65\%}   \\

        \hline
        DGCRN \textit{w / add}  & 2.65 & 5.17 & 6.76\%   & 3.00 & 6.15 & 8.18\%  & 3.42 & 7.25 & 9.76\%   \\
        
        DGCRN \textit{w / cat}  & \textbf{2.59} & \textbf{5.00} & \textbf{6.58\%}   & \textbf{2.96} & \textbf{6.05} & \textbf{7.89\%}  & \textbf{3.38} & \textbf{7.12} & \textbf{9.51\%}   \\
    
	\bottomrule
\end{tabular}}
\label{tab:add-cat}
\end{table*}

\begin{figure*}[t]
\centering
\vspace{8pt}
\begin{tabular}{c@{ }@{ }c@{ }@{ }c@{ }@{ }c}

    \begin{minipage}{1.0\linewidth}\centering
        \includegraphics[scale=0.5745]{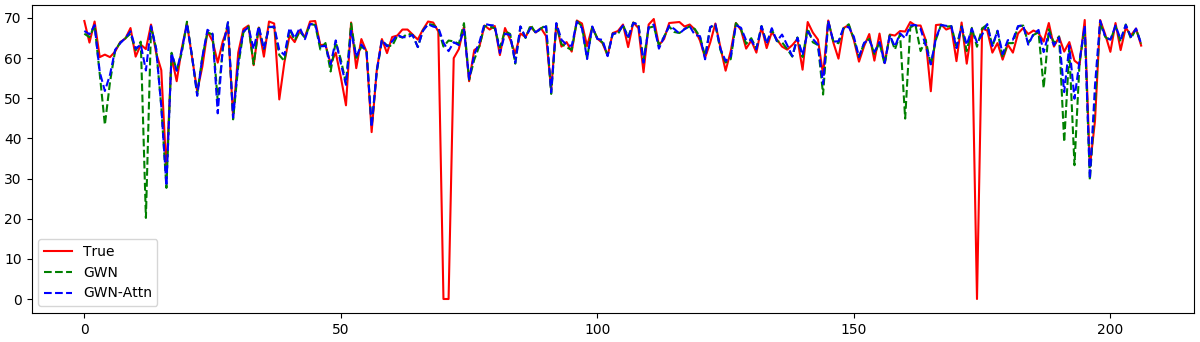}
    \end{minipage} & \vspace{12pt} \\
    

    \begin{minipage}{1.0\linewidth}\centering
        \includegraphics[scale=0.5745]{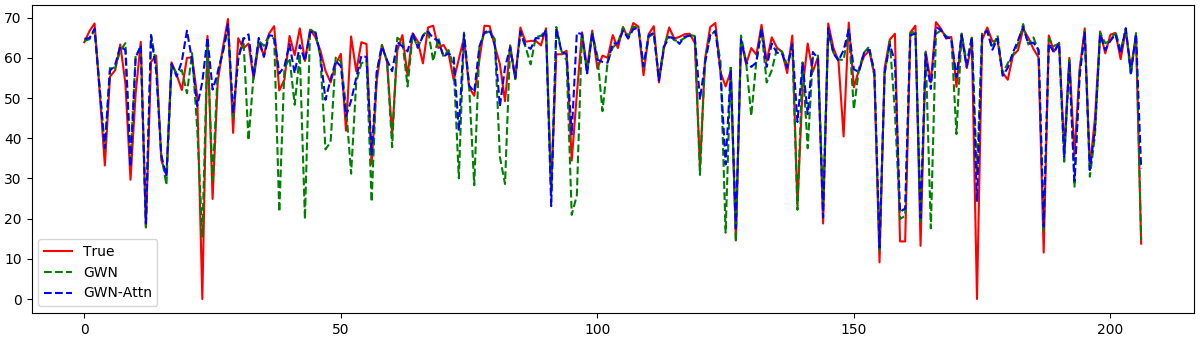}
    \end{minipage} & \vspace{12pt} \\

    \begin{minipage}{1.0\linewidth}\centering
        \includegraphics[scale=0.5745]{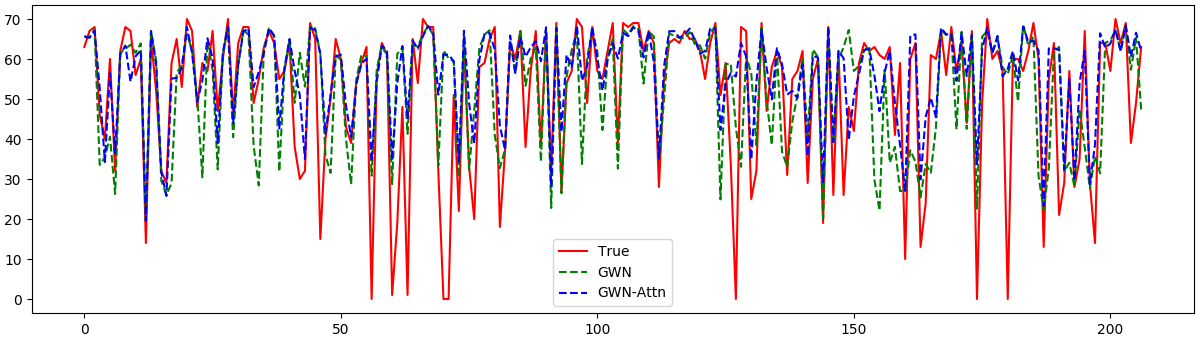}
    \end{minipage} & \\
    
\end{tabular}
\caption{The prediction performance of our method compared with GWN~\cite{graphwave2019} on METR-LA dataset. Follow the previous works~\cite{dcrnn2017}, the target zero-valued speeds are not evaluated.}
\label{fig:vis-performance}
\end{figure*}

\begin{table*}[!h]
\vspace{5pt}
\centering
\caption{Analysis of hyper-parameter $R$. The experiments are constructed with GWN~\cite{graphwave2019} on METR-LA dataset.}
\scalebox{1.05}{
	\begin{tabular}{lclll|lll|lll}  
	    \toprule
		 \multirow{2}*{Methods}  & \multirow{2}*{Number $R$}  &  \multicolumn{3}{c}{15 min}  &  \multicolumn{3}{c}{30 min}  &  \multicolumn{3}{c}{60 min} \\
		 
		\cline{3-11}
        & & MAE & RMSE & MAPE  & MAE & RMSE & MAPE & MAE & RMSE & MAPE \\

        GWN~\cite{graphwave2019}  & / & 2.71 & 5.16 & 7.11\%   & 3.11 & 6.24 & 8.54\%  & 3.56 & 7.31 & 10.10\%   \\
        
        \hline
        
        GWN \textit{w / CompFormer} & 10 & 2.69 & 5.15 & 6.85\%   & 3.05 & 6.16 & 8.33\%  & 3.48 & 7.19 & 9.96\%   \\
        
        GWN \textit{w / CompFormer} & 8 & 2.68 & 5.15 & 6.90\%  & 3.04 & 6.18 & 8.34\%  & 3.44 & 7.20 & 9.98\% \\
        
        GWN \textit{w / CompFormer} & 5   & 2.67 & 5.09 & 6.83\%  & 3.03 & 6.07 & 8.19\%  & 3.43 & 7.06 & 9.68\% \\
        
        GWN \textit{w / CompFormer} & 3   & 2.68 & 5.14 & 6.87\%  & 3.04 & 6.14 & 8.25\%  & 3.47 & 7.22 & 10.01\% \\
        
        GWN \textit{w / CompFormer} & 1 & 2.68 & 5.15 & 6.99\%  & 3.06 & 6.20 & 8.38\%  & 3.49 & 7.29 & 9.89\% \\
        
	\bottomrule
\end{tabular}}
\label{tab:gwn-aimm-la}
\end{table*}

\subsection{Evaluation Metrics}
\label{sec:metric}
Following the baselines ~\cite{dcrnn2017,mtgnn2020,graphwave2019}, we evaluate the performances of different forecasting methods by three commonly used metrics in traffic prediction:
\begin{itemize}
    \item Mean Absolute Error (MAE), which is a basic metric to reflect the actual situation of the prediction accuracy.
    \item Root Mean Squared Error (RMSE), which is more sensitive to abnormal value.
    \item  Mean Absolute Percentage Error (MAPE) which can eliminate the influence of data value magnitude  to some extent by normalization. 
\end{itemize}
The detailed formulations of these three metrics are:
\begin{align*}
    MAE (x, \hat{x}) &= \frac{1}{N} \sum_{i =1}^N \|x_i - \hat{x}_i\|, \\
    RMSE (x, \hat{x}) &= \sqrt{\frac{1}{N} \sum_{i =1}^N \|x_i - \hat{x}_i\|^2}, \\
    MAPE (x, \hat{x}) &= \frac{1}{N} \sum_{i =1}^N \frac{\|x_i - \hat{x}_i\|}{\|x_i\|},
\end{align*}
where $x = (x_1, ...,  x_N)$ denotes the ground truth of the speed of $N$ sensors, $\hat{x} = (\hat{x}_1, ...,  \hat{x}_N)$ represents the predicted values.

\subsection{Main Results}
In this section, we show the main results to demonstrate the effectiveness of our proposed framework in boosting the latest strong baselines. Specifically, we present the overall improvements on different metrics achieved by our method (Section \ref{sec:overall-performance}) and its superiority in predicting for the extreme events (Section \ref{sec:superority-extreme-events}).

\label{sec:overall-performance}
Table \ref{tab:results} summarizes the main results of our methods and the baselines on the two traffic datasets. 
For each dataset, the results can be divided into two parts: the first part is the result of the representative baselines and the second part gives the results of the six latest baselines with/without our method. 

The results show that integrated with our method, all the six latest baselines can be boosted effectively, which leads to significantly better performance than the representative baselines. For example, in predicting the speed of next 60 minutes on METR-LA, we can decrease the MAPE loss of GTS~\cite{gts2021} by 28.2\%. 

Moreover, from these results, we can see that the loss decreasing achieved in predicting the speed of next 60 minutes is much larger than those of 15 and 30 minutes. To be precise, on PEMS-BAY, we can decrease the MAPE loss of DCRNN~\cite{dcrnn2017} by up to 7.14\% in predicting the speeds of next 60 minutes. This implies that our method can achieve much more significant improvement if we predict the future for more steps, whose advantage comes from our test-time compensated representation learning framework in prediction.

\subsection{Superiority in Extreme Events}
\label{sec:superority-extreme-events}

Figure \ref{fig:input_attn_zero_entropy} present the relationship between the effect of our ComFormer module and the extremeness. We can see that the magnitude of the attention output rapidly increases when the number of zero-valued speed or the entropy of the input becomes larger. This implies that the attention module extracts much more auxiliary information from the periodic data bank to assist prediction and plays a more important role in these extreme events. 

Figure \ref{fig:loss-gap} presents the relationship among the improvement achieved by our framework, which is measured by the gained loss on GWN~\cite{graphwave2019} and DGCRN~\cite{dgcrn2021}, the effect of our modules and the extremeness. The consistent shapes of these curves demonstrate that in the events with higher degree of extremeness, our modules can achieve more significant improvement in prediction as the loss gap is much larger. The results above consistently verify that our modules can boost the baselines in face of the extreme events. 

Figure \ref{fig:vis-result-dgcrn} presents the prediction results of GWN~\cite{graphwave2019} with and without our attention module on some events with various degrees of extremeness. It shows that when the extremeness is low, the improvement achieved by our module is limited; while for the extreme events, the improvement is much more significant. As shown in Figure~\ref{fig:vis-performance}, when predicting vehicle speed on complex road conditions, our proposed modules can significantly improve prediction performance.

\subsection{Ablation Study}
In this section, we perform ablation studies to examine the impact of different settings on our proposed framework.

\subsubsection{Effect of concatenated compensation}
Our method concatenates instead of adds the spatial attention output to the original input. As shown in Table \ref{tab:add-cat}, the concatenated compensated features outperform the added approach on both MTGNN~\cite{mtgnn2020} and DGCRN~\cite{dgcrn2021}.

\subsubsection{Effect of hyper-parameter R}
The results presented in Table \ref{tab:gwn-aimm-la} indicate that our module is not sensitive to the value of $R$, and it can achieve good performance even when we sample only 5 series. This allows us to reduce the computational complexity of attention via uniform sampling, which is made possible by our well-constructed periodic data bank that utilizes spatio-temporal decomposition.

\section{Conclusion}
\label{sec:conclusion}
In this paper, we propose a framework for test-time compensated representation learning to improve the performance of existing traffic forecasting models under extreme events. The key idea is that the traffic patterns exhibit obvious periodic characteristics, thus we can store them separately in a periodic data bank and extract valuable information at a lower cost. This enables the model to learn useful information from the periodic data bank via a spatial transformer to to compensate for the unreliable recent observations during extreme events. The proposed CompFormer only needs to focus on the spatial dimension and transfer the compensated representation via a spatial attention score. The experimental results demonstrate the effectiveness of our method in improving the performance of existing strong baselines.


\bibliographystyle{abbrv}
\bibliography{refs}

\end{document}